\documentclass[twocolumn]{article}

\usepackage{graphicx} 
\usepackage{subfigure} 

\usepackage{natbib}

\usepackage{algorithm}
\usepackage{algorithmic}

\usepackage{amsmath}
\usepackage{amsfonts}
\usepackage{algorithm}
\usepackage{algorithmic}
\usepackage{graphicx,color}
\usepackage{rotating}
\usepackage{authblk}

\newcommand{\argmin}{\operatornamewithlimits{argmin}}

\title{The Potential Benefits of Filtering Versus Hyper-Parameter Optimization}

\author{Michael R. Smith\thanks{msmith@axon.cs.byu.edu}}
\author{Tony Martinez}
\author{Christophe Giraud-Carrier}
\affil{Department of Computer Science,
            Brigham Young University, Provo, UT 84602 USA}

\date{}
\begin{document} 
\maketitle

\begin{abstract}
The quality of an induced model by a learning algorithm is dependent on the quality of the training data \textit{and} the hyper-parameters supplied to the learning algorithm.
Prior work has shown that improving the quality of the training data (i.e., by removing low quality instances) or tuning the learning algorithm hyper-parameters can significantly improve the quality of an induced model.
A comparison of the two methods is lacking though.
In this paper, we estimate and compare the potential benefits of filtering and hyper-parameter optimization.
Estimating the potential benefit gives an overly optimistic estimate but also empirically demonstrates an approximation of the maximum potential benefit of each method.
We find that, while both significantly improve the induced model, improving the quality of the training set has a greater potential effect than hyper-parameter optimization.
%
\end{abstract}

\section{Introduction}
\label{s:intro}
The goal of supervised machine learning is to induce an accurate generalizing function (hypothesis) $h$ from a set of input feature vectors $X =\{x_1, x_2, \dots ,x_n\}$ and a corresponding set of of label vectors $Y =\{y_1, y_2, \dots, y_n\}$ that maps $X \mapsto Y$ given a set of training instances $T:\langle X, Y\rangle$.
The quality of the induced function $h$ by a learning algorithm is dependent on the values of the learning algorithm's hyper-parameters \textit{and} the quality of the training data $T$.
It is well-known that real-world data sets are often noisy.
It is also known that no learning algorithm or hyper-parameter setting is best for all data sets (no free lunch theorem \citep{Wolpert1996}).
Thus, it is important to consider \textit{both} the quality of the data and the learning algorithm with its associated hyper-parameters when inducing a model of the data.

Prior work has shown that hyper-parameter optimization \citep{Bergstra2012} and improving the quality of the training data (i.e., correcting \citep{Kubica2003}, weighting \citep{Rebbapragada2007}, or filtering \citep{Smith2011} suspected noisy instances) can significantly improve the generalization of an induced model.
However, searching the hyper-parameter space and improving the quality of the training data have generally been examined in isolation.
In this paper, we compare the effects of hyper-parameter optimization and improving the quality of the training data through filtering.
We estimate the potential benefit of filtering and hyper-parameter optimization by choosing the subset of training instances/hyper-parameters that produce the highest 10-fold cross-validation accuracy for each data set.
Maximizing the 10-fold cross validation accuracy, in a sense, overfits the data (although none of the data points in the validation set are used for training).
However, maximizing the 10-fold cross-validation accuracy provides more perspective on the magnitude of the potential improvement provided by each method.
The results could then be analyzed to determine algorithmically which subset of the training data and/or hyper-parameters to use for a given task and learning algorithm.

For filtering, we use an ensemble filter \citep{Brodley1999} as well as an adaptive ensemble filter.
In an ensemble filter, an instance is removed if it is misclassified by $n\%$ of the learning algorithms in the ensemble.
The adaptive ensemble filter is built by greedily searching for learning algorithms from a set of candidate learning algorithms that results in the highest cross-validation accuracy on the entire data set when it is added to the ensemble filter.
For hyper-parameter optimization, we use random hyper-parameter optimization \citep{Bergstra2012}.
Given the same amount of time constraints, the authors showed that a random search of the hyper-parameters out performs a grid search.

We find that filtering has the potential for a significantly higher increase in classification accuracy than hyper-parameter optimization.
On a set of 52 data sets using 6 learning algorithms with default hyper-parameters, the average classification accuracy increases from 79.15\% to 87.52\% when the training data is filtered.
Hyper-parameter optimization for the 6 learning algorithms only increases the average accuracy to 81.61\%.
The significant improvement in accuracy caused by filtering demonstrates the magnitude of the potential benefit that filtering could have on classification accuracy.
These results provide motivation for further research into developing algorithms that improve the quality of the training data.

\section{Preliminaries}
\label{section:prelims}
In this section, we establish the preliminaries and notation that will be used to discuss hyper-parameter optimization and the quality of the training data.
Given that in most cases, all that is known about a task is contained in the set of training instances, at least initially, the instances in a data set are generally considered equally.
Therefore, with most machine learning algorithms, one is concerned with maximizing $p(h|T)$, where $h: X \rightarrow Y$ is a hypothesis or function mapping input feature vectors $X$ to their corresponding label vectors $Y$, and $T=\{(x_i,y_i): x_i\in X \wedge y_i \in Y\}$ is a training set.
Generally, there is some overfit avoidance for a learning algorithm to maximize $p(h|T)$ while also minimizing the expected loss on validation data.
The goodness of the induced hypothesis $h$ is then characterized by its empirical error for a specified loss function $L$ on a validation set $V$: $$E(h, V) = \frac{1}{|V|} \sum_{<x_i, y_i> \in V} L(h(x_i), y_i).$$
In practice, $h$ is induced by a learning algorithm $g$ trained on $T$ with hyper-parameters $\lambda$, i.e., $h=g(T,\lambda)$.

Characterizing the success of a learning algorithm at the data set level (e.g., accuracy or precision) optimizes $p(h|T)$ over the entire training set and marginalizes the impact of a single training instance on an induced model.
Some instances can be more beneficial than other instances for inducing a model of the data and some instances can even be detrimental.
By \textit{detrimental instances}, we mean instances that have a negative impact on the model induced by a learning algorithm.
For example, outliers or mislabeled instances are not as beneficial as border instances and are detrimental in many cases.
In addition, other instances can be detrimental for inducing a model of the data even if they are labeled correctly and are not outliers.
Formally, a detrimental instance $\langle x_d,y_d\rangle$ is an instance that, when it is used for training, increases the empirical error:
$$E(g(T, \lambda),V) > E(g(T-\langle x_d,y_d \rangle,\lambda),V).$$

\begin{figure}[t]
\centering
\input{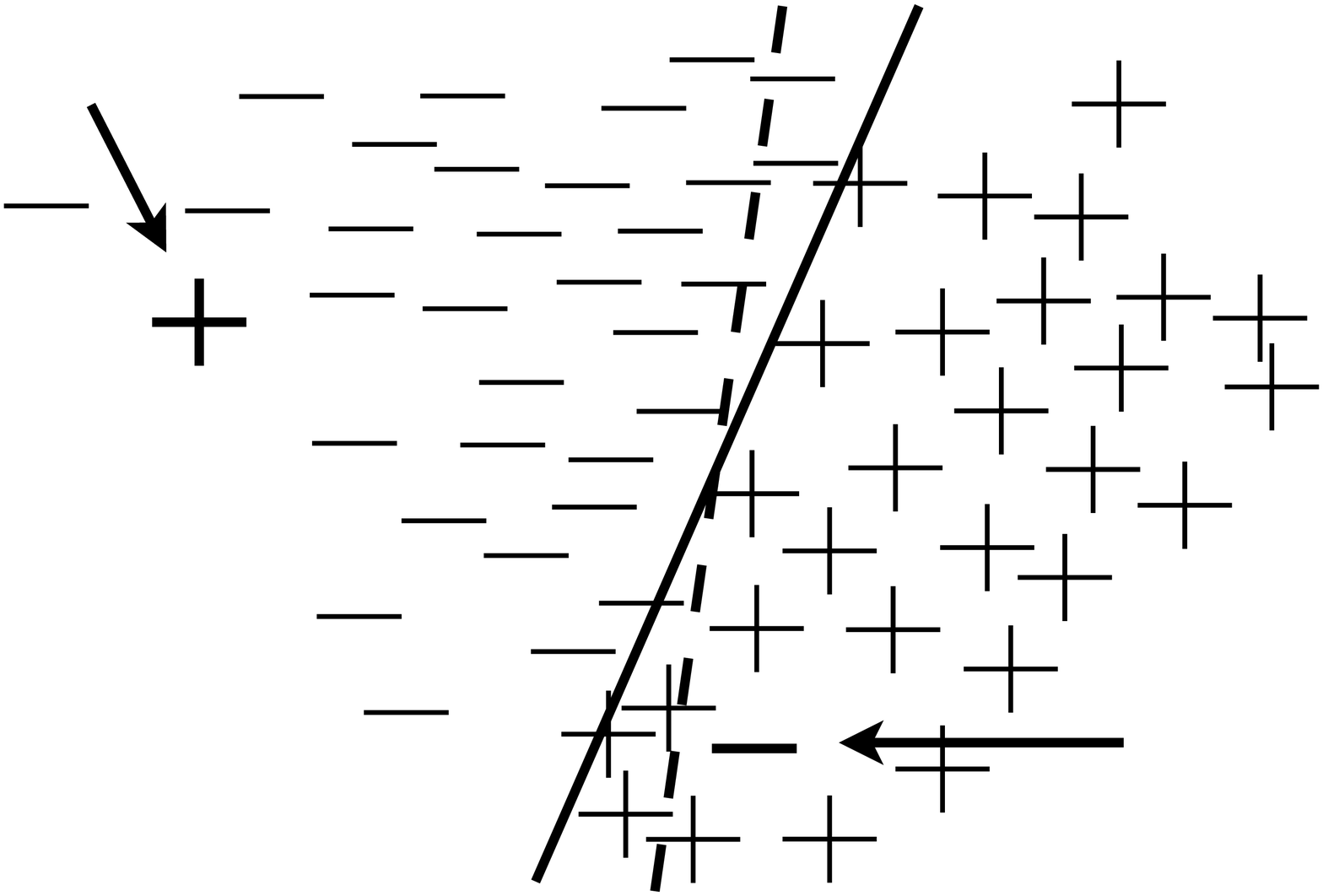} \\
\caption{Hypothetical 2-dimensional data set that shows the effects of detrimental instances in the training data on a learning algorithm.}
\label{figure:example}
\end{figure}

The effects of training with detrimental instances is demonstrated in the hypothetical two-dimensional data set shown in Figure \ref{figure:example}.
Instances A and B represent detrimental instances.
The solid line represents the ``actual'' classification boundary and the dashed line represents a potential induced classification boundary.
Instances A and B adversely affect the induced classification boundary because they ``pull'' the classification boundary and cause several other instances to be misclassified that otherwise would have been classified correctly even though the induced classification boundary (the dashed line) may maximize $p(h|T)$.

Mathematically, the effect of each instance on the induced hypothesis can be found through a decomposition of Bayes' theorem:
\begin{align}
p(h|T) &= \frac{p(T|h)p(h)}{p(T)} \nonumber\\
&= \frac{\prod_i^{|T|} p(\langle x_i, y_i\rangle|h)p(h)}{p(T)}.\label{eq:instance}
\end{align}
Despite having a mechanism to avoid overfitting detrimental instances (often denoted as $p(h)$), the presence of detrimental instances still affects the induced model for most learning algorithms.
Detrimental instances have the most significant impact during the early stages of training where it is difficult to identify detrimental instances and undo the negative effects caused by them \citep{Elman1993}.

In the sections that follow, we discuss how a) hyper-parameter optimization and b) improving the quality of the data handle detrimental instances.

\subsection{Hyper-parameter Optimization}
\label{section:hyper-paramOpt}
\begin{figure}[t]
\centering
\begin{tabular}{cc}
\input{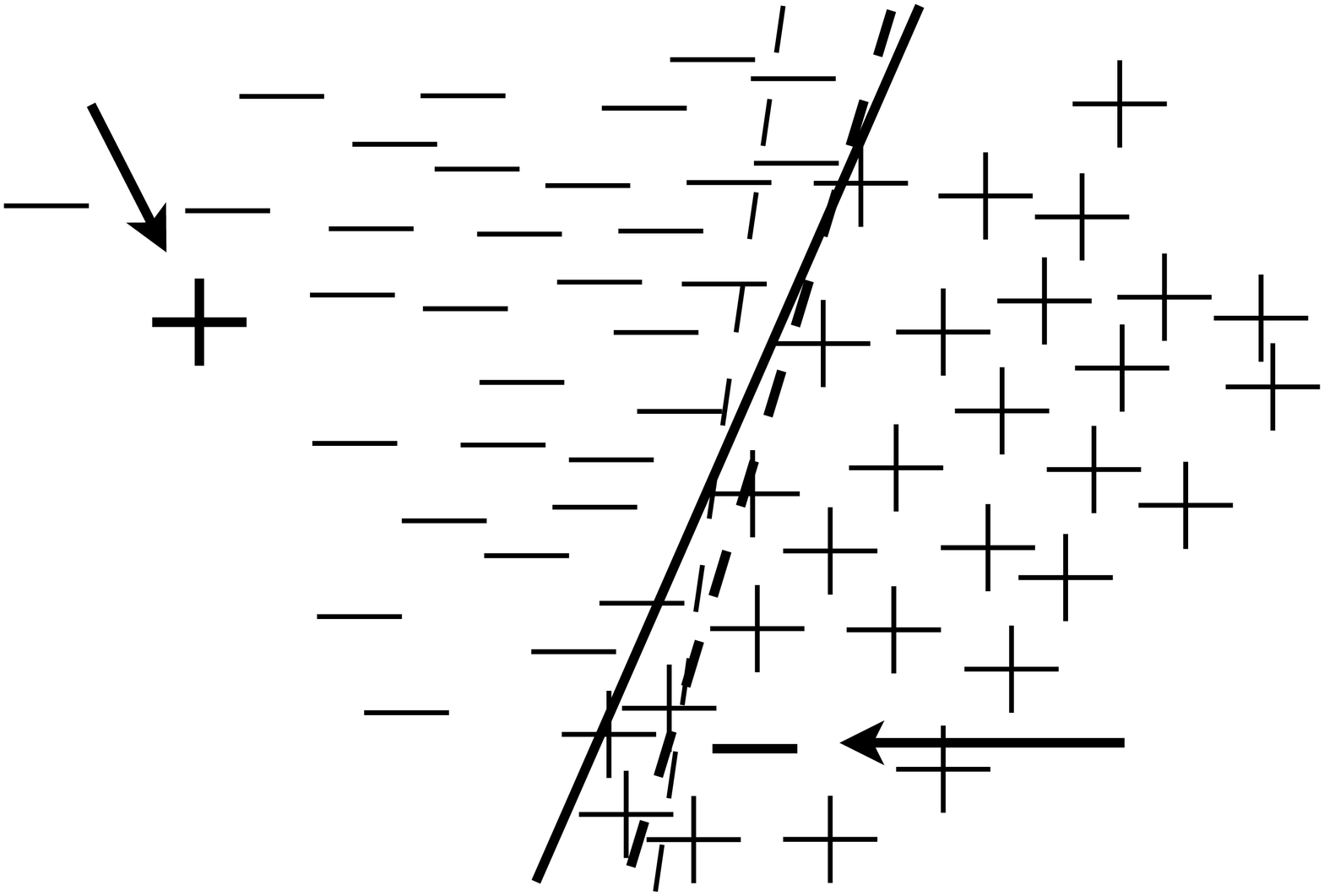} & \input{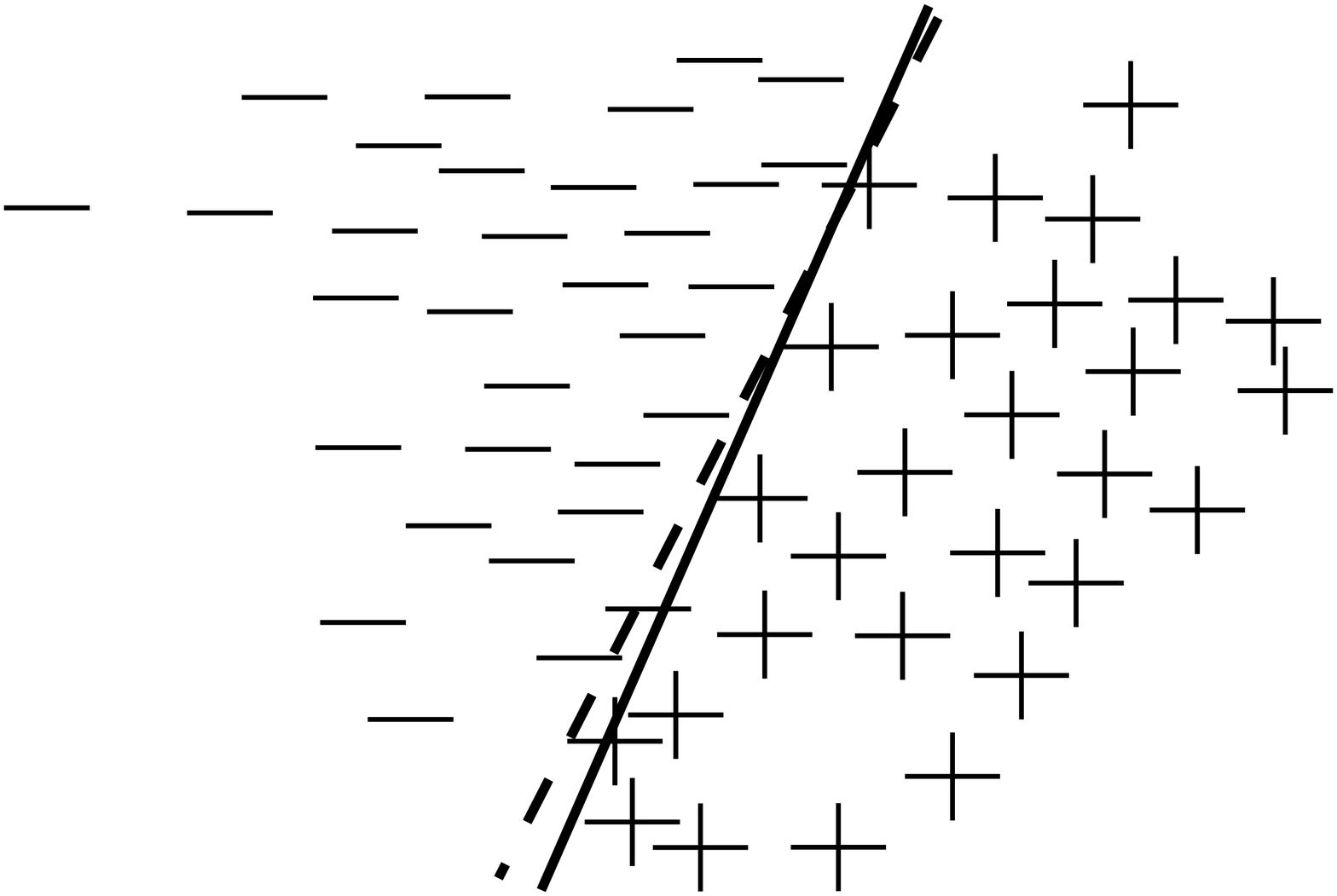} \\
a & b\\
\end{tabular}
\caption{Hypothetical data set that shows the effects of suppressing detrimental instances in the training data on a learning algorithm with a) hyper-parameter optimization and b) filtering.}
\label{figure:filter}
\end{figure}

The quality of an induced model by a learning algorithm depends in part on the learning algorithm's hyper-parameters.
With hyper-parameter optimization, the hyper-parameter space $\Lambda$ is searched to minimize the empirical error on $V$:
\begin{equation}
\argmin_{\lambda \in \Lambda} E(g(T,\lambda),V). \label{eq:hyper-paramOpt}
\end{equation}
The hyper-parameters can have a significant effect on suppressing detrimental instances and inducing better models in general.
For example, the loss function in a support vector machine could be set to use the ramp-loss function which limits the penalty on instances that are too far from the decision boundary \citep{Collobert2006}.
Suppressing the effects of the detrimental instances improves the induced model, but does not change the fact that detrimental instances still affect the model.
This is shown graphically in Figure \ref{figure:filter}a using the same hypothetical 2-dimensional data set in Figure \ref{figure:example}.
The original induced hypothesis without hyper-parameter optimization is shown as the gray dashed line.
The new induced hypothesis is represented as the bold dashed line.
The effect of instance A and instance B may be reduced but instance A and instance B still affect the induced hypothesis.
As shown mathematically in Equation \ref{eq:instance}, each instance is still considered during the learning process.


\subsection{Improving the Training Data Quality}
\label{section:dataQuality}
The quality of an induced model also depends on the quality of the training data.
Low quality training data results in lower quality induced models.
The quality of each training instance is generally not considered when searching for the most probable hypothesis given the training data other than overfit avoidance.
Thus, the learning process could also seek to improve the quality of the training data--such as searching for a subset of the training data that results in lower empirical error:
$$\argmin_{t \in \mathcal{P}(T)} E(g(t,\lambda),V)$$
where $t$ is a subset of $T$ and $\mathcal{P}(T)$ is the power set of $T$.
The effects of training on only a subset of the training data is shown in Figure \ref{figure:filter}b.
The detrimental instances A and B are removed from the data set and obviously have no effect on the induced model.
Mathematically, it is readily apparent that the removed instances have no effect on the induced model as they are not included in the product in Equation \ref{eq:instance}.

\section{Identifying Detrimental Instances}
\label{section:search}

\begin{figure*}[t]
\centerline{\input{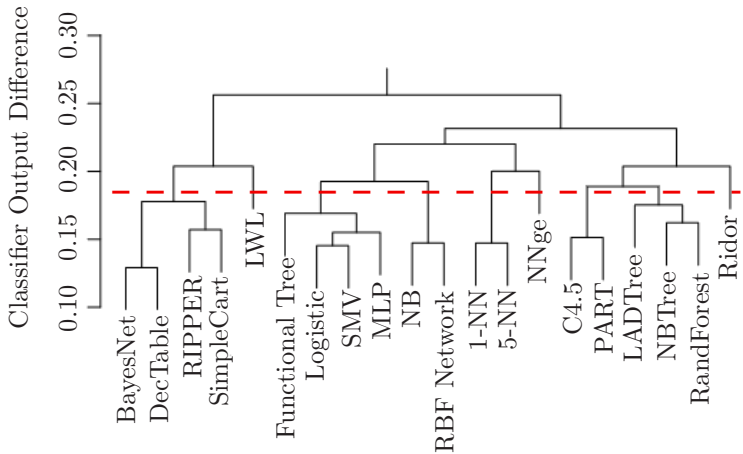}}
\caption{Dendrogram of the considered learning algorithms clustered using unsupervised metalearning based on their classifier output difference. The dashed line represents the cut-point of 0.18 COD value used to create the clusters from which the learning algorithms were selected.}
\label{figure:COD}
\end{figure*}

Recognizing that some instances are detrimental for inducing a classification function leads to the question of how to identify which instances are detrimental.
Identifying detrimental instances is a non-trivial task.
Fully searching the space of subsets of training instances generates $2^N$ subsets of training instances where $N$ is the number of training instances.
Even for small data sets, it is computationally infeasible to induce $2^N$ models to determine which instances are detrimental.
There is no known way to determine how a single instance will affect the induced classification function without inducing a classification function with the investigated instance removed from the training set.

As previously discussed, most learning algorithms induce the most probable classification function given the training data.
For classification problems, we are interested in maximizing the probability of a class label given a set of input features: $p(y_i|x_i, h)$.
Using the chain rule, $p(\langle x_i,y_i\rangle| h)$ from Equation \ref{eq:instance} can be substituted with $p(y_i|x_i, h) p(x_i|h)$:
$$
p(h|T) = \frac{\prod^{|T|}_i p(y_i|x_i, h) p(x_i|h)p(h)}{p(T)}.
$$
Each instance contributes to $p(h|T)$ as the quantity $p(y_i|x_i,h)p(x_i|h)$.
Previous work in noise handling has shown that class noise is more detrimental than attribute noise \citep{Zhu2004,Nettleton2010}.
Thus, searching for detrimental instances that have a low $p(y_i|x_i,h)$ is a natural place to start.

Recently, \citet{Smith2012_IH} investigated the presence of instances that are likely to be misclassified in commonly used data sets as well as their characteristics.
We follow their procedure of using a set of diverse learning algorithms to estimate $p(y_i|x_i,h)$.
The dependence of $(p_i|x_i, h)$ on $h$ can be lessened by summing over the space of all possible hypotheses:
\begin{equation}
p(y_i|x_i) = \sum_{h \in \mathcal{H}} p(y_i|x_i, h)p(h|T). \label{equation:IH}
\end{equation}
Practically, to sum over $\mathcal{H}$, one would have to sum over the complete set of hypotheses, or, since $h=g(T,\lambda)$, over the complete set of learning algorithms and hyper-parameters associated with each algorithm.
This, of course, is not feasible.
In practice, $p(y_i|x_i)$ can be estimated by restricting attention to a diverse set of representative algorithms (and hyper-parameters).
Also, it is important to estimate $p(h|T)$ because if all hypotheses were equally likely, then all instances would have the same $p(y_i|x_i)$ under the no free lunch theorem \citep{Wolpert1996}.
A natural way to approximate the unknown distribution $p(h|T)$ is to weight a set of representative learning algorithms, and their associated hyper-parameters, $\mathcal{L}$, a priori with an equal, non-zero probability while treating all other learning algorithms as having zero probability.
Given such a set $\mathcal{L}$ of learning algorithms, we can then approximate Equation~\ref{equation:IH} to the following:
\begin{align}
p(y_i|x_i) \approx \frac{1}{|\mathcal{L}|} \sum_{j=1}^{|\mathcal{L}|} p(y_i|x_i, g_j(T,\lambda))
\end{align}
where $p(h|T)$ is approximated as $\frac{1}{|\mathcal{L}|}$ and $g_j$ is the j$^{th}$ learning algorithm from $\mathcal{L}$.
The distribution $p(y_i|x_i,g_j(T,\lambda))$ is estimated using the indicator function with 5 by 10-fold cross-validation (running 10-fold cross validation 5 times, each time with a different seed to partition the data) since not all learning algorithms produce probabilistic outputs.

To get a good representation of $\mathcal{H}$, and hence a reasonable estimate of $p(y_i|x_i)$, we select a diverse set of learning algorithms using unsupervised metalearning (UML)~\citep{Lee2011}.
UML uses Classifier Output Difference (COD)~\citep{Peterson2005} to measure the diversity between learning algorithms.
COD measures the distance between two learning algorithms as the probability that the learning algorithms make different predictions.
UML then clusters the learning algorithms based on their COD scores with hierarchical agglomerative clustering.
Here, we considered 20 commonly used learning algorithms with their default hyper-parameters as set in Weka \citep{weka2009}.
The resulting dendrogram is shown in Figure \ref{figure:COD}, where the height of the line connecting two clusters corresponds to the distance (COD value) between them.
A cut-point of 0.18 was chosen and a representative algorithm from each cluster was used to create $\mathcal{L}$ as shown in Table \ref{table:LA}.

\begin{table}[t]
\centering
\caption{Set of learning algorithms $\mathcal{L}$ used to estimate $p(y_i|x_i)$.}
\label{table:LA}
\setlength{\tabcolsep}{3pt}
\begin{tabular}{ll}
\hline

& \multicolumn{1}{c}{Learning Algorithms}\\
\hline

*& Multilayer Perceptron trained with Back \\&Propagation (MLP) \\
*& Decision Tree (C4.5) \\
*& Locally Weighted Learning (LWL) \\
*& 5-Nearest Neighbors (5-NN) \\
*& Nearest Neighbor with generalization (NNge) \\
*& Na\"{i}ve Bayes (NB) \\
*& RIpple DOwn Rule learner (RIDOR) \\
*& Random Forest (RandForest) \\
*& Repeated Incremental Pruning to Produce Error \\
&Reduction (RIPPER) \\
\hline
\end{tabular}
\end{table}


\begin{algorithm}[tb]
  \caption{Adaptively identifying detrimental instances.}
  \label{figure:greedySearch}
\begin{algorithmic}[1]
\STATE Let $F$ be the set of learning algorithms used for filtering, $\mathcal{L}$ be the set of candidate learning algorithms for $F$, and $\phi$ be the threshold such that instances with a $p(y_i|x_i)$ less than $\phi$ are filtered from the training set. 
\STATE Initialize $F$ to the empty set: $F \gets \{\}$ 
\STATE Initialize current accuracy to the accuracy without filtering: $currAcc\gets runLA(\{\})$. $runLA(F)$ returns the cross-validation accuracy from a learning algorithm trained on a data set filtered with $F$.\label{line:init}
\WHILE{$\mathcal{L} \neq \{\}$} 
   \STATE $bestAcc \gets currAcc$; $bestLA \gets null$;
   \FORALL{$g \in \mathcal{L}$} 
       \STATE $tempF \gets F + g$; $acc\gets runLA(tempF)$;
       \IF {$acc > bestAcc$}
          \STATE $bestAcc \gets acc$; $bestLA \gets g$;
       \ENDIF
   \ENDFOR
   \IF{$bestAcc > currAcc$}
      \STATE $\mathcal{L} \gets \mathcal{L} - bestLA$; $F \gets F + bestLA$;  $currAcc \gets bestAcc$;
   \ELSE
      \STATE break;
   \ENDIF
\ENDWHILE
\end{algorithmic}
\end{algorithm}

In addition to using the entire set $\mathcal{L}$ of learning algorithms in the ensemble filter (the $\mathcal{L}$-filter), we also dynamically create the ensemble filter from $\mathcal{L}$ using a greedy algorithm.
This allows us to find a specific set of learning algorithms that are best for filtering a given data set and learning algorithm combination.
Algorithm \ref{figure:greedySearch} outlines our approach.
The adaptive ensemble filter $F$ is constructed by iteratively adding the learning algorithm $g$ from $\mathcal{L}$ that produces the highest cross-validation classification accuracy when $g$ is added to the ensemble filter.
Because we are using the probability that an instance will be misclassified rather than a binary yes/no, we also use a threshold $\phi$ to determine which instances to examine as being detrimental.
Instances with a $p(y_i|x_i)$ less than $\phi$ are discarded from the training set.
A constant threshold value for $\phi$ is set to filter the instances in $runLA(F)$ for all iterations.
The baseline accuracy for the adaptive approach is the accuracy of the learning algorithm without filtering (line \ref{line:init}).
The search stops once adding one of the remaining learning algorithms to the ensemble filter does not increase accuracy, or all of the learning algorithms in $\mathcal{L}$ have been added to the ensemble filter.

The adaptive filtering approach overfits the data since the cross validation accuracy is maximized (all detrimental instances are included for evaluation).
This allows us to find those instances that are actually detrimental in order to examine the effects that they can have on an induced model.
Of course, this is not feasible in practical settings, but provides insight into the potential improvement gained from improving the quality of the training data.

\section{Filtering Versus Hyper-Parameter Optimization}
\label{section:results}
In this section, we empirically compare the \textit{potential} effects of filtering detrimental instances with those of hyper-parameter optimization.
When comparing two learning algorithms, statistical significance is determined using the Wilcoxon signed-ranks test \citep{Demsar2006} with an alpha value of 0.05.
In addition to reporting the average classification accuracy, we also report two relative measures--the average percentage of reduction in error and the average percentage of reduction in accuracy, calculated as:
\begin{displaymath}
\begin{array}{lr}
\%reduction_{err} = \frac{g(d)-bl(d)}{100-bl(d)}, & \textnormal{if }  g(d) \ge bl(d)\\
\%reduction_{acc} = \frac{g(d)-bl(d)}{bl(d)}, & \textnormal{if } g(d) < bl(d)\\ 
\end{array}
\end{displaymath}
where $bl(d)$ returns the baseline accuracy on data set $d$ and $g(d)$ returns the accuracy on data set $d$ of the algorithm that we are interested in (i.e.,~filtering or hyper-parameter optimization).
The percent reduction in error captures the fact that increasing the accuracy from 98\% to 99\% is more difficult than increasing the accuracy from 50\% to 51\%, which is not expressed in the average accuracy.
Likewise, the percent reduction in accuracy captures the relative difference in loss of accuracy for the investigated method.
We also show the number of times that $g(d)$ is greater than, equal to, or less than $bl(d)$ (\textit{count}).
Together, the count, the percent reduction in error, and the percent reduction in accuracy show the potential benefit of the using an algorithm and also the variance of an algorithm.
For example, an algorithm has more variance than another if it increases the percent reduction in error but also increases the percent reduction in accuracy compared to an algorithm that may not increase the percent reduction in error as much but also does not increase the percent reduction in accuracy.
The average accuracy for each algorithm on a data set is determined using 5 by 10-fold cross-validation.

Hyper-parameter optimization uses 10 random searches of the hyper-parameter space.
Random hyper-parameter selection was chosen based on the work by \citep{Bergstra2012}.
The premise of random hyper-parameter optimization is that most machine learning algorithms have very few hyper-parameters that considerably affect the final model while most of the other hyper-parameters have little to no effect on the final model.
Random search provides a greater variety of the hyper-parameters that considerably affect the model, thus allowing for better selection of these hyper-parameters.
Given the same amount of time constraints, random search has been shown to out perform a grid search.
For reproducibility, the exact process of hyper-parameter optimization for the learning algorithms is provided in the supplementary material.
The accuracy from the hyper-parameters that resulted in the highest cross-validation accuracy for each data set is reported.

For filtering using the ensemble ($\mathcal{L}$-filter) and adaptive filtering, we use thresholds $\phi$ of 0.5, 0.3, and 0.1 to identify detrimental instances.
The $\mathcal{L}$-filter estimates $p(y_i|x_i)$ using all of the learning algorithms in the set $\mathcal{L}$ (Table \ref{table:LA}).
The adaptive filter greedily constructs an ensemble to estimate $p(y_i|x_i)$ for a specific data set/learning algorithm combination.
The accuracy from the $\phi$ that produced the highest accuracy for each data set is reported.

To show the effect of filtering detrimental instances and hyper-parameter optimization on an induced model, we examine filtering and hyper-parameter optimization in six commonly used learning algorithms: MLP, C4.5, IBk, NB, Random Forest, and RIPPER.
The LWL, NNge, and Ridor learning algorithms are not used for analysis because they do not scale well with the larger data sets--not finishing due to memory overflow or large amounts of running time with many hyper-parameter settings.\footnote{For the data sets on which the learning algorithms did finish, the effects of hyper-parameter optimization and filtering on LWL, NNge, and Ridor are consistent with the other learning algorithms.}

\begin{table}
\centering
\caption{A comparison of the effects of an ensemble filter and of hyper-parameter optimization on the performance of a learning algorithm.}
\setlength{\tabcolsep}{2.15pt}
\vskip 0.15in
\begin{tabular}{l c c c c c c  }
\hline

 & MLP & C4.5 & IB\textit{k} & NB & RF & RIP \\
\hline

Orig & 80.74 & 80.11 & 79.03 & 75.68 & 81.59 & 77.83 \\
\hline

$\mathcal{L}$-filter & \textbf{83.53} & \textbf{82.57} & \textbf{82.90} & \textbf{78.51} & \textbf{83.79} & \textbf{81.37} \\
\%red$_{\text{err}}$ & 16.01 & 12.43 & 18.28 & 14.92 & 14.65 & 14.96 \\
\%red$_{\text{acc}}$ & -0.64 & -1.23 & -1.08 & -0.47 & -0.70 & -0.64 \\

count & 44,1,7 & 45,1,6 & 44,2,6 & 42,0,10 & 38,3,11 & 50,0,2 \\
\hline

HPO & \textbf{83.14} & \textbf{81.93} & \textbf{82.15} & \textbf{79.63} & \textbf{82.75} & \textbf{80.04} \\
\%red$_{\text{err}}$ & 20.24 & 12.73 & 20.02 & 22.77 & 19.81 & 12.59 \\
\%red$_{\text{acc}}$ & -2.83 & -1.11 & -0.32 & -0.67 & -2.13 & -0.48 \\

count & 47,0,5 & 39,0,13 & 41,2,9 & 42,1,9 & 37,2,13 & 47,1,4 \\
\hline
\end{tabular}
\label{table:paramOptVsOrig}
\vskip -0.1in
\end{table}

The results comparing the accuracy of the default hyper-parameters set in weka \citep{weka2009} with the $\mathcal{L}$-filter and hyper-parameter optimization (HPO) are shown in Table \ref{table:paramOptVsOrig}.
The increases in accuracy that are statistically significant are in bold.
Not surprisingly, both the $\mathcal{L}$-filter and hyper-parameter optimization significantly increase the classification accuracy for all of the investigated learning algorithms.
The magnitude of the increase in average accuracy is similar for both approaches.
However, hyper-parameter optimization shows more variance than filtering as demonstrated by larger percent reduction in error \textit{and} in a larger percent reduction in accuracy.
The $\mathcal{L}$-filter also generally increases the accuracy on more data sets than hyper-parameter optimization.

\begin{table}[t]
\centering
\caption{A comparison of the effects of filtering with hyper-parameter optimization on the performance of a learning algorithm.}
\setlength{\tabcolsep}{1.85pt}
\vskip 0.15in
\begin{tabular}{l cccccc}
\hline

 & MLP & C4.5 & IB\textit{k} & NB & RF & RIP \\
\hline

HPO & 83.14 & 81.93 & 82.15 & 79.63 & 82.75 & 80.04 \\
\hline

$\mathcal{L}$-F$_\text{orig}$ & 83.53 & \textbf{82.57} & 82.90 & 78.51 & \textbf{83.79} & \textbf{81.37} \\
\%red$_\text{err}$ & 10.31 & 9.43 & 8.26 & 13.14 & 12.33 & 11.98 \\
\%red$_\text{acc}$ & -2.11 & -2.45 & -1.94 & -5.65 & -1.64 & -2.57 \\

count & \begin{small}27,3,22\end{small} & \begin{small}33,4,15\end{small} & \begin{small}30,2,20\end{small} & \begin{small}22,2,28\end{small} & \begin{small}27,1,24\end{small} & \begin{small}34,1,17\end{small} \\
\hline

$\mathcal{L}$-F$_\text{HPO}$ & \textbf{84.50} & \textbf{83.08} & \textbf{83.83} & \textbf{81.05} & \textbf{85.22} & \textbf{82.85} \\
\%red$_\text{err}$ & 11.26 & 9.24 & 9.39 & 8.33 & 12.34 & 12.60 \\
\%red$_\text{acc}$ & -0.39 & -0.13 & -0.89 & -0.39 & -0.06 & -0.20 \\

count & 35,4,13 & 39,5,8 & 43,2,7 & 43,2,7 & 45,1,6 & 45,5,2 \\
\hline

A$_\text{orig}$ & \textbf{88.24} & \textbf{87.39} & \textbf{90.08} & \textbf{81.91} & \textbf{90.17} & \textbf{87.34} \\
\%red$_{\text{err}}$ & 37.87 & 35.22 & 48.41 & 22.93 & 47.02 & 37.98 \\
\%red$_{\text{acc}}$ & -1.41 & -0.47 & -0.41 & -4.69 & -1.47 & -1.42 \\

count & 45,0,7 & 48,2,2 & 51,0,1 & 34,0,18 & 46,0,6 & 48,0,4 \\
\hline

A$_\text{HPO}$ & \textbf{85.78} & \textbf{84.45} & \textbf{84.87} & \textbf{82.25} & \textbf{86.57} & \textbf{84.24} \\
\%red$_{\text{err}}$ & 19.23 & 17.67 & 16.26 & 14.66 & 22.67 & 21.43 \\
\%red$_{\text{acc}}$ & 0.00 & 0.00 & -0.90 & -0.23 & 0.00 & NA \\

count & 50,1,1 & 47,3,2 & 49,1,2 & 49,0,3 & 50,1,1 & 50,1,0 \\
\hline
\end{tabular}
\label{table:paramOptVsFiltering}
\vskip -0.1in
\end{table}

Table \ref{table:paramOptVsFiltering} compares the hyper-parameter optimized learning algorithms with filtering.
$\mathcal{L}$-F$_\text{orig}$ and $\mathcal{L}$-F$_\text{HPO}$ represent using the $\mathcal{L}$-filter where the learning algorithms in $\mathcal{L}$ have default hyper-parameter settings and where the hyper-parameters of the learning algorithms in $\mathcal{L}$ have been optimized.
Likewise, A$_\text{HPO}$ and A$_\text{orig}$ represent the results from the adaptive filtering algorithm when the ensemble filter is built from $\mathcal{L}$ with and without hyper-parameter optimization.
Compared with hyper-parameter optimization, the $\mathcal{L}$-filter without hyper-parameter optimization significantly increases the classification accuracy for the C4.5, random forest and RIPPER learning algorithms.
However, if the $\mathcal{L}$-filter is composed of hyper-parameter optimized learning algorithms, then the increase in accuracy is significant for all of the considered learning algorithms.
The accuracy increases on more data sets for all of the examined learning algorithms when $\mathcal{L}$ is composed of hyper-parameter optimized learning algorithms.
Thus, in combination, hyper-parameter optimization can result in more significant gains in classification accuracy and exhibits less variance.

The adaptive filter also significantly improves classification accuracy over hyper-parameter optimization and provides a greater increase in accuracy than the $\mathcal{L}$-filter with hyper-parameter optimization.
The results from the adaptive filters show the potential gain in filtering if a filtering algorithm could more accurately identify detrimental instances for each data set and learning algorithm combination.
With and without hyper-parameter optimization, the adaptive filter results in significant gains in generalization accuracy for each learning algorithm.~These results show the impact of choosing an appropriate subset of the training data and provide motivation for improving filtering algorithms.
Building an adaptive filter from the set of learning algorithms without hyper-parameter optimization provides higher average classification accuracy but also exhibits more variance.
However, the adaptive filter composed of hyper-parameter optimized learning algorithms increases the accuracy on more data sets than the adaptive filter composed of learning algorithms with their default hyper-parameters for the MLP, na\"{i}ve Bayes, random forest, and RIPPER learning algorithms.

\begin{table}[t]
\centering
\caption{The frequency of selecting a learning algorithm when adaptively constructing a filter set.
Each row gives the percentage of cases that the learning algorithm was included in the filter set for the learning algorithm in the column.}
\setlength{\tabcolsep}{1.5pt}
\vskip 0.15in
\begin{tabular}{l c c c cccc}
\hline

 & ALL & MLP & C4.5 & IB5 & NB & RF & RIP \\
\hline

None & 5.36 & 2.69 & 2.95 & 3.08 & 5.64 & 5.77 & 1.60 \\
MLP & 18.33 & 16.67 & 15.77 & 20.00 & \textbf{25.26} & 23.72 & 16.36 \\
C4.5 & 17.17 & 17.82 & 15.26 & 22.82 & 14.49 & 13.33 & 20.74 \\
IB5 & 12.59 & 11.92 & 14.23 & 1.28 & 10.00 & 17.18 & 16.89 \\
LWL & 6.12 & 3.59 & 3.85 & 4.36 & 23.72 & 3.33 & 3.59 \\
NB & 7.84 & 5.77 & 6.54 & 8.08 & 5.13 & 10.26 & 4.92 \\
NNge & 19.49 & \textbf{26.67} & 21.15 & 21.03 & 11.15 & \textbf{24.74} & 23.40 \\
RF & \textbf{21.14} & 22.95 & \textbf{26.54} & \textbf{23.33} & 15.77 & 15.13 & \textbf{24.20} \\
Rid & 14.69 & 14.87 & 16.79 & 18.33 & 11.92 & 16.54 & 12.77 \\

RIP & 8.89 & 7.82 & 7.69 & 8.85 & 13.08 & 7.44 & 4.39 \\
\hline
\end{tabular}
\label{table:ds_freq}
\vskip -0.1in
\end{table}

There is no one learning algorithm that is the optimal filter for all learning algorithms and/or data sets.
Table \ref{table:ds_freq} shows the frequency for which a learning algorithm with default hyper-parameters was selected for filtering by the adaptive filter.
The greatest percentage of cases a learning algorithm is selected for filtering for each learning algorithm is in bold.
The column ``ALL'' refers to the average from all of the learning algorithms as the base learner.
No instances are filtered in 5.36\% of the cases.
Thus, filtering to some extent increases the classification accuracy in about 95\% of the cases.
Furthermore, the random forest, NNge, MLP, and C4.5 learning algorithms are the most commonly chosen algorithms for filtering.
However, no one learning algorithm is selected in more than 27\% of the cases.
The filtering algorithm that is most appropriate is dependent on the data set and the learning algorithm.
Previously, \citet{Saez2013} examined when filtering is most beneficial for the nearest neighbor classifier.
They found that the efficacy of noise filtering is dependent on the characteristics of the data set and provided a rule set to determine when filtering will be most beneficial.
Future work includes better understanding the efficacy of noise filtering for each learning algorithm and determining \textit{which} filtering approach to use for a given data set.

\section{Related Work}
\label{section:relatedWorks}
To our knowledge, this is the first work that examines \textit{both} filtering and hyper-parameter optimization in handling detrimental instances and compares their effects.
Our work was motivated in part by \citet{Smith2012_IH} who examined the existence of and characterization of instances that are hard to classify correctly.
They found that a significant number of instances are hard to classify correctly and that the hardness of each instance is dependent on its relationship with the other instances in the training set as well as the learning algorithm used to induce a model of the data.
Thus, there is a need for improving the way detrimental instances are handled during training as they affect the classification of the other instances in the data set.

Other previous work has examined improving the quality of the training data and hyper-parameter optimization in isolation.
\citet{Frenay2014} provide a survey of the current approaches for dealing with label noise in classification problems which often take the approach of improving the quality of the data.
Improving the quality of the training data has typically fallen into three approaches: filtering, cleaning, and instance weighting.
Each technique within an approach differs in how detrimental instances are identified.
A common technique for filtering removes instances from a data set that are misclassified by a learning algorithm \citep{Tomek1976} or an ensemble of learning algorithms \citep{Brodley1999}.
Removing the training instances that are suspected to be noise and/or outliers prior to training has the advantage that they do not influence the induced model and generally increase classification accuracy \citep{Smith2011}.
Training on a subset of the training data has also been observed to increase accuracy in active learning \citep{Schohn2000}.

A negative side-effect of filtering is that beneficial instances can also be discarded and produce a worse model than if all of the training data had been used \citep{Smith_Eval}.
Rather than discarding the instances from a training set, one approach seeks to clean or correct instances that are noisy or possibly corrupted  \citep{Kubica2003}.
However, this could artificially corrupt valid instances.
Weighting, on the other hand, weights suspected detrimental instances rather than discarding them \citep{Rebbapragada2007,Smith_RIDL}.
Weighting the instances allows for an instance to be considered on a continuum of detrimentality rather than making a binary decision of an instance's detrimentality.
These approaches have the advantage that data is not discarded which is especially beneficial when data is scarce.

Much of the previous work has artificially inserted noise and/or corrupted instances into the initial data set to determine how well an approach would work in the presence of noisy or mislabeled instances.
In some cases, a given approach only has a significant impact when there are large degrees of artificial noise.
We show that correctly labeled, non-noisy instances can {\em also} be detrimental for inducing a model of the data and that properly handling them can result in significant gains in classification accuracy.
Thus, in contrast to much previous work, we did not artificially corrupt a data set to create detrimental instances.
Rather, we sought to identify the detrimental instances that are already contained in a data set.

The grid search and manual search are the most common types of hyper-parameter optimization techniques in machine learning.
A combination of the two approaches are commonly used \citep{Larochelle2007}.
Bayesian optimization has also been used to search the hyper-parameter space \citep{Snoek2012, Hutter2011} as an alternative approach.
Further, \citet{Bergstra2012} proposed to use a random search of the hyper-parameter space as discussed previously.

\section{Conclusions}
\label{section:conclusions}
In this paper we compared the potential benefits of filtering with hyper-parameter optimization both mathematically and empirically.
Mathematically, hyper-parameter optimization may reduce the effects of detrimental instances on an induced model but the detrimental instances are still considered in the learning process.
Filtering, on the other hand, fully removes the detrimental instances--completely eliminating the effects of the detrimental instances on the induced model.

Empirically, we estimated the potential benefits of each method by maximizing the 10-fold cross-validation accuracy of each data set.
For the adaptive filter, we also chose a set of learning algorithms for an ensemble filter by maximizing the 10-fold cross-validation accuracy.
Both filtering and hyper-parameter optimization significantly increase the classification accuracy for all of the considered learning algorithms.
On average, a learning algorithm increases in classification accuracy from 79.15\% to 87.52\% by removing the detrimental instances on the observed data sets using the adaptive filter.
On the other hand, hyper-parameter optimization only increases the average classification accuracy to 81.61\%.
Filtering with hyper-parameter optimized learning algorithms produces more stable results (i.e. consistent increases in classification accuracy and low decreases in classification accuracy) than filtering or hyper-parameter optimization in isolation.

One of the reasons that instance subset selection is overlooked is due to the fact that there is a large computational requirement to observe the relationship of an instance on the other instances in a data set.
As such, determining which instances to filter is non-trivial.
We hope that the results presented in this paper will provide motivation for furthering the work in improving the quality of the training data.


%
%

\bibliographystyle{icml2014}%
\bibliography{../../bibliography}

\label{lastpage}

\end{document}